\documentclass[11pt]{article}
\pdfoutput=1
\usepackage{EMNLP2023}

\usepackage{times}
\usepackage{latexsym}
\usepackage[T1]{fontenc}
\usepackage[utf8]{inputenc}
\usepackage{inconsolata}
\usepackage{microtype}
\usepackage{graphicx}
\usepackage{subcaption}
\usepackage{booktabs}
\usepackage{amsmath}
\usepackage{amssymb}
\usepackage{mathtools}
\usepackage{amsthm}
\usepackage{enumitem}
\usepackage{colortbl}
\usepackage{multirow}
\usepackage{siunitx}
\usepackage{tabularx}
\usepackage{ragged2e}
\usepackage{float}
\usepackage[table,xcdraw]{xcolor}
\usepackage{algorithm}
\usepackage{algpseudocode}
\usepackage{listings}
\usepackage{soul}
\usepackage[most]{tcolorbox}
\usepackage{tikz}
\usetikzlibrary{tikzmark}
\usepackage{subcaption}

\theoremstyle{plain}

\theoremstyle{definition}

\theoremstyle{remark}

\usepackage[textsize=tiny]{todonotes}

\definecolor{queryanchorcolor}{HTML}{39A432}
\definecolor{awakencolor}{HTML}{7A52A6}
\definecolor{matchcolor}{HTML}{EB8E47}
\definecolor{filtercolor}{HTML}{428DBF}

\definecolor{lightorg1}{HTML}{D9AE7B}
\definecolor{lightorg2}{HTML}{E2BF8F}
\definecolor{lightorg3}{HTML}{EED0AB}
\definecolor{lightorg4}{HTML}{FAE2C5}
\definecolor{lightorg5}{HTML}{FDEEDD}
\definecolor{lightorg6}{HTML}{FEF6EC}
\definecolor{lightorg7}{HTML}{FFFCF7}

\definecolor{lightgrey}{HTML}{D3D3D3}
\definecolor{lightblue}{HTML}{BFDCE7}
\definecolor{lightpurple}{RGB}{236,232,243}
\definecolor{lightgreen}{RGB}{231,245,235}

\title{PonsRAG: A Pons-Inspired RAG Bridging Cognitive Islands for\\Coordinated Long Narrative Reasoning}

\author{
  \textbf{Rongchen Zhao}\textsuperscript{$1^{*}$,$2$,\textdagger},
  \textbf{Yu Chen}\textsuperscript{$2$,\textdagger},
  \textbf{Juyuan Wang}\textsuperscript{$2$},
  \textbf{Zhouting Mo}\textsuperscript{$4$}
  \\
  \textbf{Jianxing Yu}\textsuperscript{$3$},
  \textbf{Wenqing Chen}\textsuperscript{$1$},
  \textbf{Jingping Liu}\textsuperscript{$1$,\textdaggerdbl}
  \\   
  \textsuperscript{$1$}School of Software Engineering, Sun Yat-sen University\\
  \textsuperscript{$2$}School of Future Technology, South China University of Technology\\
  \textsuperscript{$3$}School of Artificial Intelligence, Sun Yat-sen University\\
  \textsuperscript{$4$}TikTok Inc\\
  {\small
    \texttt{\{edwinzhaorc, cyu94987\}@gmail.com,}
    \texttt{liujp68@mail.sysu.edu.cn}
  }
}
\begin{document}
\maketitle
\begingroup
\renewcommand{\thefootnote}{}
\footnote{\textsuperscript{*}Intern. \textsuperscript{\textdagger}Equal contribution. \textsuperscript{\textdaggerdbl}Corresponding author.}
\addtocounter{footnote}{-1}
\endgroup
\begin{abstract}
Long Narrative Reasoning is an essential capability for processing and reasoning over complex narratives. While retrieval-augmented generation provides a promising framework, existing methods still face two critical challenges: cognitive islanding and cross-layer evidence disconnection. To address these issues, we propose PonsRAG, a coordinated RAG framework inspired by the biological pons. PonsRAG consists of two key components: Triple-Layer Indexing, which organizes documents into a connected knowledge structure to bridge cognitive islands, and Coordinated Reasoning, which retrieves evidence across distinct layers and integrates cross-layer information into a unified context. We evaluate PonsRAG on four long-context narrative benchmarks, and experimental results show that it outperforms the strongest baseline, achieving a 11.56\% relative improvement in average accuracy on multi-choice tasks.
\end{abstract}
\section{Introduction}
Long Narrative Reasoning (LNR) refers to the ability of models to process and reason over extended narratives, maintaining context across multiple characters and plots. Unlike multi-hop tasks \cite{zhou2025disentangling}, which connect distant evidence across diverse documents, LNR requires synthesizing information from long and complex texts, making it essential for applications such as advanced dialogue systems, content summarization, and story generation.
\begin{figure}[t]
\centering
\includegraphics[width= 1\columnwidth]{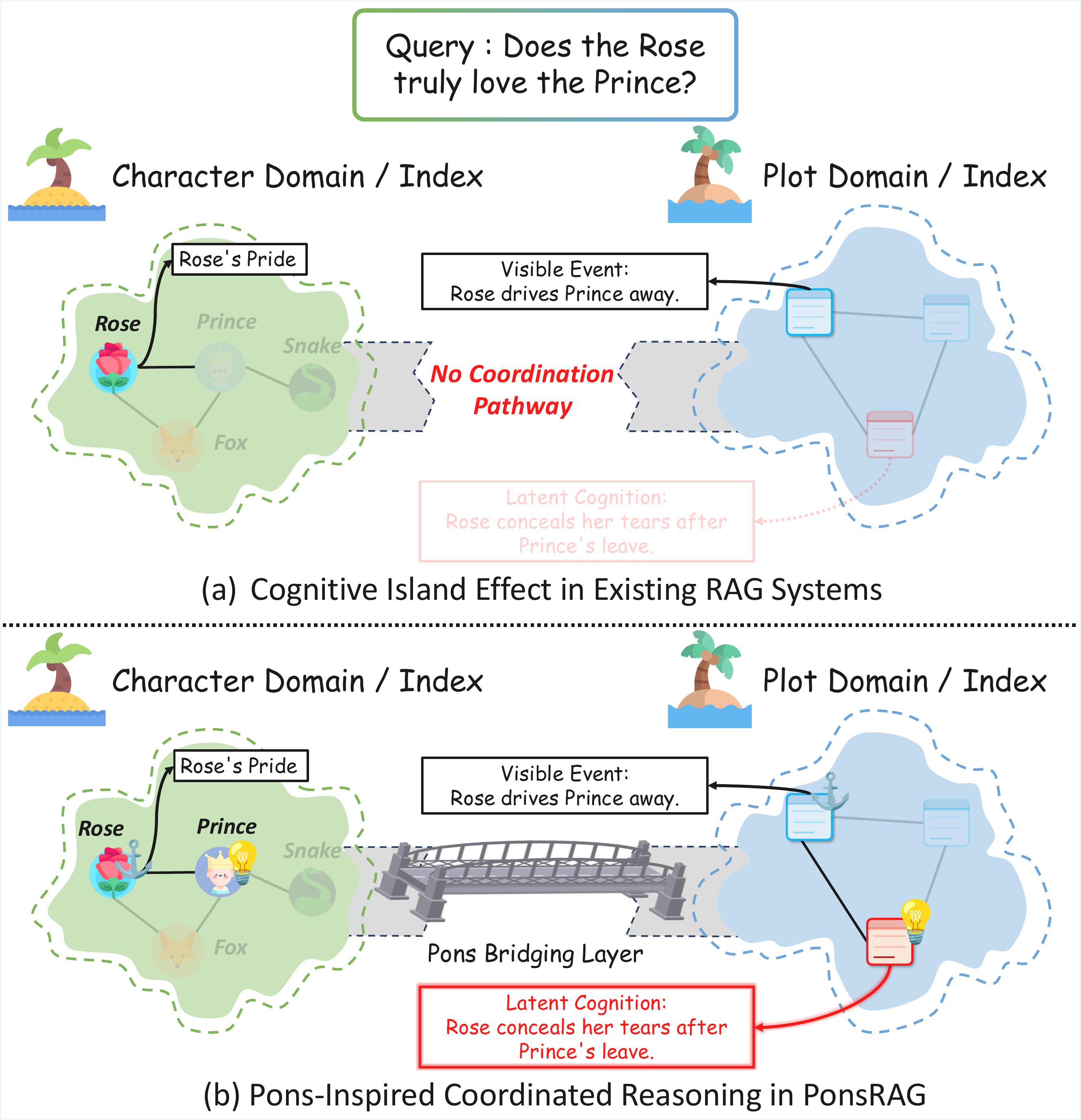}
\caption{Bridging cognitive islands in long narrative reasoning.}
\label{fig:subfig}
\vspace{-1.5ex}
\end{figure}
Retrieval-Augmented Generation (RAG) has emerged as a promising solution for LNR, which builds a retrieval index over the story to retrieve query-relevant evidence for reasoning. Based on their indexing mechanism, recent RAGs can be categorized into two paradigms: Single-Layer and Multi-Layer Indexing. Frameworks adopting \textbf{Single-Layer Indexing} organize chunks of the document into a structured knowledge layer as their retrieval index to augment LNR. Much effort has been dedicated to designing various knowledge layers, including HippoRAGv2~\citep{pmlr-v267-gutierrez25a} and RAPTOR~\citep{sarthi2024raptor}. However, Single-Layer Indexing faces the challenge that relying on a single knowledge layer provides only a partial view of the evidence. For instance, given the query \textit{“Does the Rose truly love the Prince?”} (The Little Prince), a knowledge graph may capture the relationship between the Rose and the Prince but fail to capture higher-level events, such as \textit{the Rose's rejection}, leading to failure. In contrast, \textbf{Multi-Layer Indexing} presents a more advantageous solution. It captures distinct views of the evidence by establishing separate knowledge layers. Many studies along this line have demonstrated its effectiveness such as the factual-semantic-episodic index \citep{wang2025comorag}, the community hierarchical index and the community graph-tree index \citep{dong2025youtugraphrag}. They significantly improve reasoning by providing multi-view context retrieved from distinct layers.
Thus, in this work, we adopt the multi-layer indexing paradigm to support LNR. However, existing methods still face two practical challenges. First, during the offline index construction stage, evidence stored in different knowledge layers is in isolation, making it difficult to connect semantically related information across layers. We use the term Cognitive Island to describe this cross-layer evidence disconnection: character-centric and plot-centric evidence may be individually relevant, yet remain separated during retrieval. For example, the traits and actions of Rose are closely related but may reside in different layers, as illustrated in Figure \ref{fig:subfig}(a). Second, during the online reasoning stage, retrieval is typically conducted independently within each layer, with limited coordination across layers. As a result, the retrieved context may be fragmented, redundant, or even locally biased, making it harder for the model to assemble a coherent reasoning chain. In the example of Figure \ref{fig:subfig}(b), answering the query requires linking evidence across layers, such as \textit{Rose $\rightarrow$ Prince $\rightarrow$ Rose's rejection $\rightarrow$ Rose's tears}.
To address these two issues, we propose PonsRAG, a RAG framework inspired by Pons, a neural architecture. It plays an important relay role in coordinating information flow across distributed brain regions \citep{kandel2013principles, fernandez2010anatomy, palesi2017contralateral}. By linking otherwise separated neural areas through dense fiber pathways, it supports the integration of signals required for complex tasks \citep{manto2012consensus, kratochwil2017long,zhang2026towards}. Based on this theory, PonsRAG introduces a Triple-Layer Indexing architecture with Character, Plot, and Pons layers. The central Pons Layer is implemented as a bipartite graph that connects nodes across distinct layers, enabling cross-layer evidence propagation and selection. Built on this index, PonsRAG further employs a Coordinated Reasoning pipeline that retrieves, matches, and filters evidence jointly across layers, transforming retrieval from isolated layer-wise search into coordinated cross-layer evidence construction.
Our primary contributions are summarized as follows:
\begin{itemize}[noitemsep,nolistsep,leftmargin=*]
\item {We build on the bridge-and-relay concept of the biological pons for LNR, where PonsRAG bridges separated evidence across different knowledge layers, facilitating the integration of cross-layer information.}
\item {We propose PonsRAG, a triple-layer retrieval index paired with a coordinated reasoning pipeline. Its key contribution is shifting from the isolated retrieval of characters and plots to a bridge-constrained joint selection and link fragmented evidence to enable cross-layer reasoning.}
\item {We evaluate PonsRAG on four long-context narrative benchmarks and it achieves the best performance across all benchmarks against all baselines, improving the relative average accuracy by 11.56\% on Multi-Choice tasks.}
\end{itemize}
\begin{figure*}[t]
\centering
\includegraphics[width=2\columnwidth]{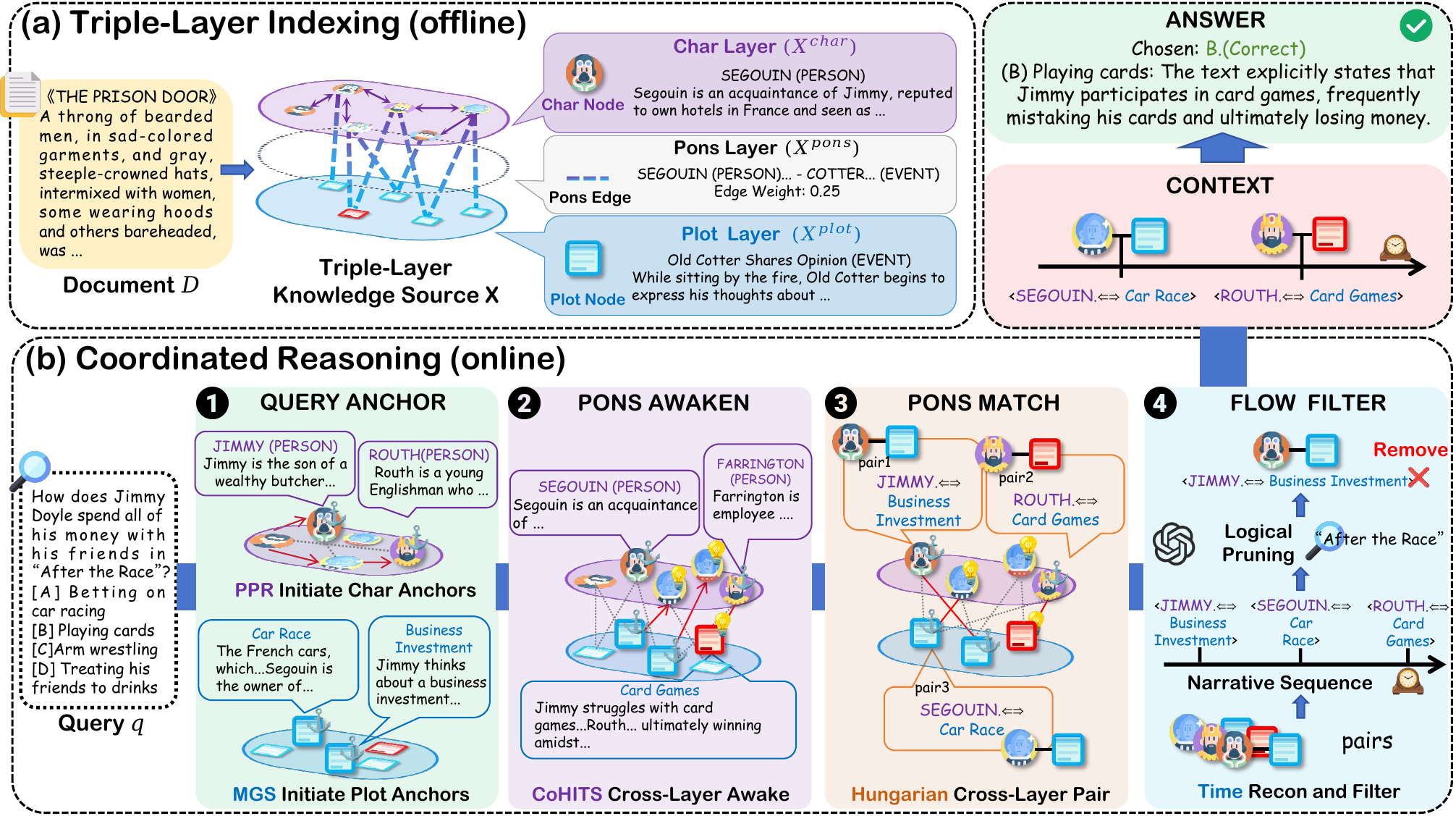}
\caption{Overall architecture of PonsRAG.}
\label{fig:mainfig}
\vspace{-1.5ex}
\end{figure*}
\section{Related Work}
We classify recent RAGs based on their indexing mechanism into two categories: Single-Layer Index RAGs and Multi-Layer Index RAGs.
\subsection{Single-Layer Index RAGs} 
Existing research into Single-Layer Index RAG typically rely on a monolithic retrieval index to facilitate knowledge acquisition \citep{chen2023end}. For instance, RAPTOR \citep{sarthi2024raptor} recursively clusters chunks to build a semantic summary tree which effectively captures events at varying levels of granularity. HippoRAGv2 \citep{pmlr-v267-gutierrez25a} focuses on relationships between entities by constructing an entity-centric graph over documents and adopts Personalized PageRank (PPR) \citep{haveliwala2002topic} to retrieve evidence based on the query relative entity. GraphRAG \citep{microsoft2024graphrag} constructs a knowledge graph index over document-level entities and relations, and augments retrieval by combining local graph evidence with community-level summaries distilled from the graph.
\subsection{Multi-Layer Index RAGs}
Multi-layered RAG frameworks transcend flat indexing by organizing knowledge into different knowledge layers for retrieval\cite{zhang2026balancing}. ComoRAG \citep{wang2025comorag} constructs a triple-layer index over veridical, semantic, and episodic knowledge from document, and retrieves evidence from each layer according to a predefined proportion. HiRAG \citep{huang2025retrieval} constructs multiple knowledge layers from local to global and retrieves evidence by locating high-level relevant contexts and refining to fine-grained evidence. Youtu-GraphRAG \citep{dong2025youtugraphrag} employs a schema-guided agent to construct a four-layer knowledge tree and retrieves evidence by decomposing complex queries into schema-aligned sub-queries for retrieval. 
\section{Overview}
In this section, we formalize the problem definition of the long narrative reasoning task and outline our proposed framework to tackle this problem.
\subsection{Problem Formulation} 
Formally, given a long narrative document $\mathcal{D}$ (usually exceeding $200\text{k}$ tokens) and a specific query $q$, our objective is to generate the optimal answer $A$. This task is modeled as maximizing the conditional probability:
\begin{equation}
    \hat{A} = \operatorname*{argmax}_{A} P(A \mid \mathcal{D}, q)
\end{equation}
\subsection{Our Framework} 
As illustrated in Figure~\ref{fig:mainfig}, we decouple the framework into two stages:
\paragraph{Triple-Layer Indexing (offline).} The framework begins by constructing a knowledge source $\mathcal{X}$ to serve as the index for the document $\mathcal{D}$, viewed from two complementary perspectives. The first layer, Char Layer ($\mathcal{X}^{char}$), models the traits of characters in the narrative context. The second layer, Plot Layer ($\mathcal{X}^{plot}$), captures the narrative progression, consisting of multiple atomic events. To bridge these cognitively isolated layers, we introduce the Pons Layer ($\mathcal{X}^{pons}$), analogous to the biological pons in the neural system. This layer interconnects the character and plot layers, enabling coordinated reasoning across the distributed cognition.
\paragraph{Coordinated Reasoning (online).} Building on $\mathcal{X}$, the framework implements a reasoning mechanism that mimics pons-cortex communication. Given a query, coordinated reasoning pipeline obtains the context from $\mathcal{X}$ through four steps: Query Anchor first retrieves the initial matching cognitive nodes from both $\mathcal{X}^{char}$ and $\mathcal{X}^{plot}$, respectively; Pons Awaken then leverages these anchors to discover latent cognitive nodes across layers via $\mathcal{X}^{pons}$; In the Pons Match step, aligned cognition pairs (char, plot) are formed by matching nodes from different layers;  Flow Filter reconstructs the chronological plotline from the disordered pairs, eliminates irrelevant noise, and outputs the final plotline as the context. Finally, the framework inputs the final narrative sequence with the query into the Generator to generate the optimal answer $\hat{A}$.
\section{Methodology}
In this section, we detail the two stages of our framework: Triple-Layer Indexing and Coordinated Reasoning.
\subsection{Triple Layer Indexing}\label{sec:index}
\paragraph{Char Layer: Centering Character Traits} 
Given a long narrative document $\mathcal{D}$, we construct a Character Layer $\mathcal{X}^{char}$ from a character-centric cognition perspective. We first partition $\mathcal{D}$ into a set of chunks, $C = \{c_i\}_{i=1}^N$, where $N$ denotes the number of chunks. For each chunk $c_i \in C$, we prompt an LLM to extract entities $\mathcal{E}_i = \{e_{ij}\}_{j=1}^{M_i}$, where $M_i$ denotes the number of extracted entities in chunk $c_i$, using a pre-defined character-centric schema. To capture the semantic background of each entity, we further instruct the LLM to generate a textual description $d_{ij}$ for each $e_{ij} \in \mathcal{E}_i$. A character node is then defined as $v_{ij}^{char} = (e_{ij}, d_{ij})$, and the set of character nodes extracted from chunk $c_i$ is denoted as $\mathcal{V}_i^{char}$. The complete character node set across all chunks is obtained by $\mathcal{V}^{char}=\bigcup\limits_{i=1}^N \mathcal{V}_i^{char}$. To improve retrieval recall, we also instruct the LLM to produce knowledge triples (subject-predicate-object) for each entity $e$ associated with a character node $v^{char}=(e, d_e) \in \mathcal{V}^{char}$, thereby forming a character-centric knowledge graph. These triples are integrated with the character nodes to construct the final Character Layer $\mathcal{X}^{char}$, following a strategy shown to be effective in HippoRAGv2.
\paragraph{Plot Layer: Capturing Plot Progression} 
To model the progression of the narrative, we construct a Plot Layer $\mathcal{X}^{plot}$ based on event-centric cognition along the plotline. For each chunk $c_i \in C$, we record its global position using $t_i$ and use an LLM to extract a set of discrete narrative events $\mathcal{R}_i = \{r_{ij}\}_{j=1}^{K_i}$, where $K_i$ denotes the number of extracted events in chunk $c_i$. Each event $r_{ij} \in \mathcal{R}_i$ is also assigned an event-type label $y_{ij}$. To capture high-level context, all extracted events $\mathcal{R} = \bigcup\limits_{i=1}^N \mathcal{R}_i$ are then grouped by their labels into clusters $\Phi_y = \{ r \in \mathcal{R} \mid y_r = y \}$. For each cluster $\Phi_y$, we use an LLM to generate a global cluster summary $S_y = \text{LLM}_{sum}(\Phi_y)$ and assign this summary to all events within the cluster, i.e., $s_r \leftarrow S_y$ for all $r \in \Phi_y$
Finally, inspired by Zettelkasten\citep{ahrens2017smartnotes}, we encapsulate these multi-granular representations into a Memory Card structure. A plot node is defined as $v^{plot}=(r,y_r,s_r,c_i,t_i)$, where $r$ is an event extracted from chunk $c_i$ at global position $t_i$. The resulting plot node set $\mathcal{V}^{plot}=\{v^{plot}\}$ constitutes the Plot Layer $\mathcal{X}^{plot}$.
\paragraph{Pons Layer: Bridging Cognitive Islands} 
To establish coordinated reasoning across $\mathcal{X}^{char}$ and $\mathcal{X}^{plot}$, we construct a Pons Layer $\mathcal{X}^{pons}$ that connects character nodes and plot nodes through a weighted Pons edge set $\mathcal{E}^{pons}$, defined as
\begin{equation} \label{eq:epons}
\mathcal{E}^{pons} =
\{\, (u, v, w_{uv}) \mid u \in \mathcal{V}^{char},\ v \in \mathcal{V}^{plot} \,\}.
\end{equation}
The edge weight $w_{uv}$ represents the relevance between a character node $u$ and a plot node $v$, and integrates two components. (1) Semantic relevance: a normalized semantic similarity $sim(d_u, r_v)$ between the character description $d_u$ and the event $r_v$, where the sparsity controller $\tau$ is defined to remove weak associations; (2) Frequency balancing: an inverse-frequency term $\mathcal{I}_u$, inspired by inverse document frequency (IDF), to mitigate popularity bias favoring high-frequency characters, defined as
\begin{equation} \label{eq:icf}
    \mathcal{I}_u = \log \left( \frac{N}{1 + \text{freq}(u)} \right),
\end{equation}
where $N$ denotes the total number of chunks and $freq(u)$ denotes the number of chunks in which character $u$ appears.
The final Pons edge weight $w_{uv}$ is defined as:
\begin{equation} \label{eq:wvu}
    w_{uv} = 
    \begin{cases} 
        sim(d_u, r_v) \cdot \mathcal{I}_u, &  sim(d_u, r_v) \geq \tau \\
        0, & \text{otherwise}
    \end{cases}
\end{equation}
\subsection{Coordinated Reasoning}\label{sec:retrieval}
\paragraph{Query Anchor: Query-Driven Initialization} 
To initiate coordinated reasoning across the Character Layer $\mathcal{X}^{char}$ and the Plot Layer $\mathcal{X}^{plot}$, we first anchor the query $q$ to a set of seed nodes in both layers.
For the Character Layer, we employ Personalized PageRank (PPR) over the character-centric knowledge graph. By performing query-personalized random walks, PPR assigns relevance scores to character nodes $u \in \mathcal{V}^{char}$. The top-$k_1$ ranked character nodes are selected to form the character anchor set $\mathcal{V}_{anc}^{char}$. For the Plot Layer, we design a Multi-Granularity Scoring (MGS) function to evaluate the relevance between the query $q$ and each plot node $v \in \mathcal{V}^{plot}$. The score jointly considers three complementary views: the narrative event $r_v$, its corresponding cluster summary $s_v$, and the original chunk $c_v$ from which the event is extracted. The scoring function is defined as:
\begin{equation} \label{eq:scr}
S(q, v) = \alpha \cdot sim(q, r_v) + \beta \cdot sim(q, s_v) + \gamma \cdot sim(q, c_v)
\end{equation}
where $sim(\cdot,\cdot)$ denotes cosine similarity, and $\alpha, \beta, \gamma$ are hyperparameters controlling the contribution of event-level, summary-level, and chunk-level semantics, respectively. Based on this score, the top-$k_2$ plot nodes are selected to form the plot anchor set $\mathcal{V}_{anc}^{plot}$. Consequently, the final anchor set $\mathcal{V}_{anc}$ is defined as the union of anchors from both layers:
\begin{equation} \label{eq:vanc}
\mathcal{V}_{anc} = \mathcal{V}_{anc}^{char} \cup \mathcal{V}_{anc}^{plot}.
\end{equation}
\paragraph{Pons Awaken: Cross-Layer Awakening} 
While the anchor set $\mathcal{V}_{anc}$ provides query-aware surface cognition through semantic retrieval, it remains insufficient for the deeper integration required by coordinated reasoning. Analogous to the biological pons, which serves as a relay and coordination hub for integrating signals across neural pathways, this phase leverages the Pons Layer $\mathcal{X}^{pons}$ to facilitate mutual reinforcement between the Character Layer $\mathcal{X}^{char}$ and the Plot Layer $\mathcal{X}^{plot}$. Specifically, we treat $\mathcal{V}_{anc}$ as query-activated seed signals and propagate relevance from these initial nodes across layers via the Pons connections. 
To simulate this process, we adopt the Co-HITS ranking algorithm\citep{deng2009cohits}, which establishes a bidirectional reinforcement loop between character nodes $u \in \mathcal{V}^{char}$ and plot nodes $v \in \mathcal{V}^{plot}$, through the weighted Pons Layer. Formally, this cross-layer resonance process is modeled as:
\begin{equation}
    \label{eq:resonance}
    \mathbf{p}^* = \mathcal{H}(\mathcal{V}_{anc}, \mathbf{W}), \mathbf{W}_{uv} = w_{uv},
\end{equation}
where $\mathcal{H}(\cdot)$ denotes the Co-HITS propagation operator yielding the steady-state activation $\mathbf{p}^*$, and $\mathbf{W}$ is the weighted adjacency matrix whose entries are given by the Pons edge weights $w_{uv}$ defined in the Pons Layer. 
Based on the steady-state ranking $\mathbf{p}^*$, we further identify a set of awaken nodes $\mathcal{V}_{awk}$, defined as nodes that achieve high activation scores but are not included in the initial anchor set:
\begin{equation}\label{eq:awk}
    \mathcal{V}_{awk} = \{ x \mid x \in \text{Top}_{k_3}(\mathbf{p}^*) \land x \notin \mathcal{V}_{anc} \}
\end{equation} 
Finally, we construct a candidate subgraph: $\mathcal{G}_{sub} = (\mathcal{V}_{cand}, \mathcal{E}^{sub})$ where $\mathcal{V}_{cand} = \mathcal{V}_{anc} \cup \mathcal{V}_{awk}$ and $\mathcal{E}^{sub} = \{ (u, v, w_{uv}) \in \mathcal{E}^{pons} \mid u, v \in \mathcal{V}_{cand}\}$.
\paragraph{Pons Match: Optimal Cross-Layer Pairing}
After obtaining the candidate subgraph $\mathcal{G}_{sub}$, we further prune it to derive explicit alignment pairs between the Character Layer and the Plot Layer.
Given $\mathcal{G}_{sub}$, we formulate the alignment between character nodes $u \in U=\mathcal{V}^{char} \cap \mathcal{V}_{cand}$ and plot nodes $v \in V=\mathcal{V}^{plot} \cap \mathcal{V}_{cand}$ as a Maximum Weight Bipartite Matching problem, where edge weights are defined by the Pons relevance scores $w_{uv}$. Intuitively, this formulation aims to identify a globally optimal set of character-plot pairs such that each node participates in at most one alignment while the total cross-layer relevance is maximized. To solve this problem, we employ the Hungarian Algorithm \citep{kuhn1955hungarian} to compute the optimal matching, retaining the top-$k_4$ pairs as the final matching set $\mathcal{P}_{match}$, defined as:
\begin{equation}
\begin{aligned}
&~~~~~~~~~~~~\mathcal{P}_{match}
=
\operatorname*{argmax}_{\mathcal{P} \subseteq (U,V)}
\sum_{(u,v)\in\mathcal{P}} w_{uv}, \\
&\forall (u_1,v_1),(u_2,v_2)\in\mathcal{P},
\; u_1 \neq u_2 \land v_1 \neq v_2 .
\end{aligned}
\end{equation}
The resulting matching set $\mathcal{P}_{match}$ provides a consistent alignment of the cross-layer interactions activated in the Pons Awaken phase. We restrict Pons Match to a 1-to-1 information bottleneck to isolate the evidence backbone and filter many-to-many narrative noise (see Section \ref{sec:island} for 1-to-$N$ relaxations).
\paragraph{Flow Filter: Query-Aware Answer Grounding}
\begin{table*}[!t]
\centering
\scriptsize
\setlength{\tabcolsep}{3pt}
\resizebox{1.0\textwidth}{!}{
\begin{tabular}{lcccccc|ccc}
\toprule
\textbf{Method} & \multicolumn{2}{c}{\textbf{NarrativeQA}} & \multicolumn{2}{c}{\textbf{EN.QA}} & \textbf{EN.MC} & \textbf{NoCha} & \multicolumn{2}{c}{\textbf{QA Avg.}} & \textbf{MC Avg.} \\
\cmidrule(lr){2-3} \cmidrule(lr){4-5} \cmidrule(lr){6-6} \cmidrule(lr){7-7} \cmidrule(lr){8-9} \cmidrule(lr){10-10} 
& F1 & EM & F1 & EM & ACC & ACC & F1 & EM & ACC \\
\midrule
\multicolumn{10}{c}{\cellcolor{lightgrey}\textit{\textbf{LLM}}} \\
GPT-4o-mini          & 27.29 & 7.00  & 29.83 & 12.82 & 30.57 & 60.32 & 28.56 & 9.91  & 45.45 \\
\multicolumn{10}{c}{\cellcolor{lightgrey}\textit{\textbf{Naive RAG}}} \\
BGE-M3(0.3B)         & 23.16 & 15.10 & 23.71 & 16.24 & 59.82 & 56.35 & 23.44 & 15.67 & 58.09 \\
NV-Embed-v2 (7B)     & 27.18 & \underline{17.80} & \underline{34.34} & 24.57 & 61.13 & \underline{68.25} & 30.76 & 21.19 & 64.69 \\
Qwen3-Embed-8B       & 24.19 & 15.60 & 25.79 & 17.95 & 65.50 & 57.14 & 24.99 & 16.78 & 61.32 \\
\multicolumn{10}{c}{\cellcolor{lightgrey}\textit{\textbf{Structured RAG}}} \\
RAPTOR               & 27.84 & 17.80 & 26.33 & 19.65 & 57.21 & 53.17 & 27.09 & 18.73 & 55.19 \\
HippoRAGv2           & 23.12 & 15.20 & 24.45 & 17.09 & 60.26 & 67.46 & 23.79 & 16.15 & 63.86\\

Youtu-GraphRAG       & 27.45 & 15.40 & 32.03 & 22.79 & 68.55 & 65.87 & 29.74 & 19.09 & 67.21 \\
ComoRAG\textbf{(one step)} & \underline{29.95} & 17.60 & 34.03 & \underline{24.79} & \underline{70.31} & 61.90 & 31.99 & 21.20 & 66.11 \\
\textbf{PonsRAG (Ours)} & \textbf{31.19} & \textbf{19.00} & \textbf{35.13} & \textbf{26.21} & \textbf{77.73} & \textbf{72.22} & \textbf{33.16} & \textbf{22.61} &  \textbf{74.98} \\
\midrule
\textit{\textbf{Improv.}}& \textbf{+4.14$\%$} & \textbf{+6.74$\%$} & \textbf{+2.30$\%$} & \textbf{+5.73$\%$} & \textbf{+10.55$\%$} & \textbf{+5.82$\%$} & \textbf{+3.66$\%$} & \textbf{+6.65 $\%$} & \textbf{+11.56$\%$} \\
\bottomrule
\vspace{-1.5ex}
\end{tabular}
}
\caption{\textbf{Single-step QA performance} on four long narrative comprehension datasets. For fair comparison, we adopt GPT-4o-mini as the LLM backbone. For one-step setting, we limit ComoRAG for max one step. We highlight the \textbf{best} and \underline{second-best} results. \textbf{Improv.} denotes the relative improvement of our method over the second-best baseline.}
\label{tab:single}
\end{table*}

\begin{table*}[!t]
\centering
\scriptsize
\setlength{\tabcolsep}{3pt}
\resizebox{1.0\textwidth}{!}{
\begin{tabular}{lcccccc|ccc}
\toprule
\textbf{Method} & \multicolumn{2}{c}{\textbf{NarrativeQA}} & \multicolumn{2}{c}{\textbf{EN.QA}} & \textbf{EN.MC} & \textbf{NoCha} & \multicolumn{2}{c}{\textbf{QA Avg.}} & \textbf{MC Avg.} \\
\cmidrule(lr){2-3} \cmidrule(lr){4-5} \cmidrule(lr){6-6} \cmidrule(lr){7-7} \cmidrule(lr){8-9} \cmidrule(lr){10-10} 
& F1 & EM & F1 & EM & ACC & ACC & F1 & EM & ACC \\
\midrule

\textbf{IRCoT}+RAPTOR       & 31.35 & 16.00 & 32.09 & 19.36 & 63.76 & 57.94 & 31.72 & 17.68 & 60.85 \\
\textbf{IRCoT}+HippoRAGv2   & 28.98 & 13.00 & 29.27 & 18.24 & 64.19 & 61.90 & 29.13 & 15.62 & 63.05 \\

\textbf{IRCoT}+Youtu-GraphRAG & 25.92 & 15.40 & 31.10 & 22.79 & 66.38 & \underline{63.49} & 28.51 & 19.10 & 64.94 \\
ComoRAG\textbf{(five steps)}  & \underline{31.43} & \underline{18.60} & \underline{34.52} & \underline{25.07} & \underline{72.93} & 61.90 & \underline{32.98} & \underline{21.84} & \underline{67.42} \\

IRCoT+PonsRAG (Ours) & \textbf{33.28} & \textbf{20.60} & \textbf{36.30} & \textbf{26.78} & \textbf{79.48} & \textbf{72.22} & \textbf{34.79} & \textbf{23.69} & \textbf{75.85} \\
\midrule
\textit{\textbf{Improv.}} & \textbf{+5.89\%} & \textbf{+10.75\%} & \textbf{+5.16\%} & \textbf{+6.82\%} & \textbf{+8.98\%} & \textbf{+13.75\%} & \textbf{+5.49\%} & \textbf{+8.47\%} & \textbf{+12.51\%} \\
\bottomrule
\vspace{-1.5ex}
\end{tabular}
}
\caption{\textbf{Multi-step QA performance} on four long-narrative comprehension datasets. For fair comparison, we use IRCoT for all enhanced RAGs except ComoRAG, which follows its original multi-step setting with max five steps (details in Section \ref{sec:exp_setting}).}
\label{tab:multi}
\end{table*}
The matching set $\mathcal{P}_{match}$ consists of unordered character-plot pairs and therefore does not explicitly reflect the temporal progression of the narrative. To recover narrative coherence, we first reorder the matched pairs according to the plot timestamps $t_v$ of their associated plot nodes, producing a chronologically ordered sequence $\mathcal{S}_{match}$.
Subsequently, to ensure that the retrieved context satisfies both the semantic intent and the temporal constraints implied by the query $q$, we introduce an LLM-based filtering module $\pi_{filter}$. This module selectively prunes query-irrelevant pairs and resolves temporal references expressed in the query, e.g., ``at last'' or ``earlier'', to identify the most relevant subsequence within $\mathcal{S}_{match}$. The final narrative sequence is obtained as $\mathcal{S}_{final} = \pi_{filter}(q, \mathcal{S}_{match})$.
Finally, we provide the query $q$ with the filtered narrative sequence $\mathcal{S}_{final}$ as input to the LLM, which generates the final answer $A$.
\section{Experiment}
In this section, we present the evaluation results of PonsRAG. We further conduct ablation studies and analytical experiments of our framework.
\subsection{Experimental Setup}\label{sec:exp_setting}
\paragraph{Benchmarks}
We conduct experiments on four long context narrative comprehension datasets, spanning both Question Answering (QA) and Multiple Choice (MC) tasks, including NarrativeQA \citep{kocisky2018narrativeqa}, $\infty$BENCH \citep{zhang-etal-2024-bench} (EN.QA and EN.MC) and NoCha \citep{nocha-2024-karp-thai-et-al} detailed in Appendix \ref{sec:implementation details}:
\begin{itemize}[leftmargin=*,itemsep=0pt,topsep=2pt,parsep=0pt,partopsep=0pt]
\item \textbf{NarrativeQA}: A QA dataset comprising books and movie scripts (avg. 58k tokens). Following prior work, we evaluate on a random sample of 500 test questions for computational efficiency.
\item \textbf{EN.QA}: A QA task from $\infty$BENCH containing 351 questions on classic novels, with context lengths exceeding 200k tokens.
\item \textbf{EN.MC}: An MC task from $\infty$BENCH consisting of 229 questions on classic novels, sharing similar context lengths with EN.QA.
\item \textbf{NoCha}: An MC dataset comprising 126 True/False verification questions derived from four public classic novels.
\end{itemize} 
\paragraph{Metrics}
Following previous work of ComoRAG \citep{wang2025comorag} we adopt the F1 score and Exact Match (EM) as evaluation metrics for QA tasks and utilizing Accuracy (ACC) for MC tasks.
\paragraph{Baselines}
We compare PonsRAG against three baseline categories: \textbf{(1) LLM}, which directly process the entire document context (up to 128k tokens). \textbf{(2) Naive RAG}, retrieving from flattened 512-token chunks using different embedding models, including BGE-M3~\citep{chen2024bgem3}, NV-Embed-v2~\citep{lee2024nvembed}, and Qwen3-Embed-8B~\citep{zhang2025qwen3}. \textbf{(3) Structured RAG}, which constructs structure retrieval index, including single-layer index such as RAPTOR and HippoRAGv2, and multi-layer index such as Youtu-GraphRAG and ComoRAG. We separately compare our method against GraphRAG and HiRAG in Appendix~\ref{sec:appendix_excluded_baselines}. 
\paragraph{Implementation Details}
For Single-Step QA, all RAGs execute only one single retrieval iteration with GPT-4o-mini as the LLM backbone with context length capped to 6k tokens. For fair comparison, we restrict the Meta Control Loop in ComoRAG to max one round. For Multi-Step QA, we apply IRCoT~\citep{trivedi2023ircot}, which interleaves Chain-of-Thought reasoning with iterative retrieval for those structured RAGs without native multi-step mechanisms. For ComoRAG, we retain its native multi-step setting: a max 5-round Meta Control Loop, which has been proven to be better suited for its index structure than IRCoT in LNR~\citep{wang2025comorag}. For fairness, the total context length across all steps is capped to 6k tokens. We provide the further details about experimental settings and hyperparameters of PonsRAG in Appendix \ref{sec:Hyperparameters}.
\begin{figure}[t]
\centering
\includegraphics[width=1\columnwidth]{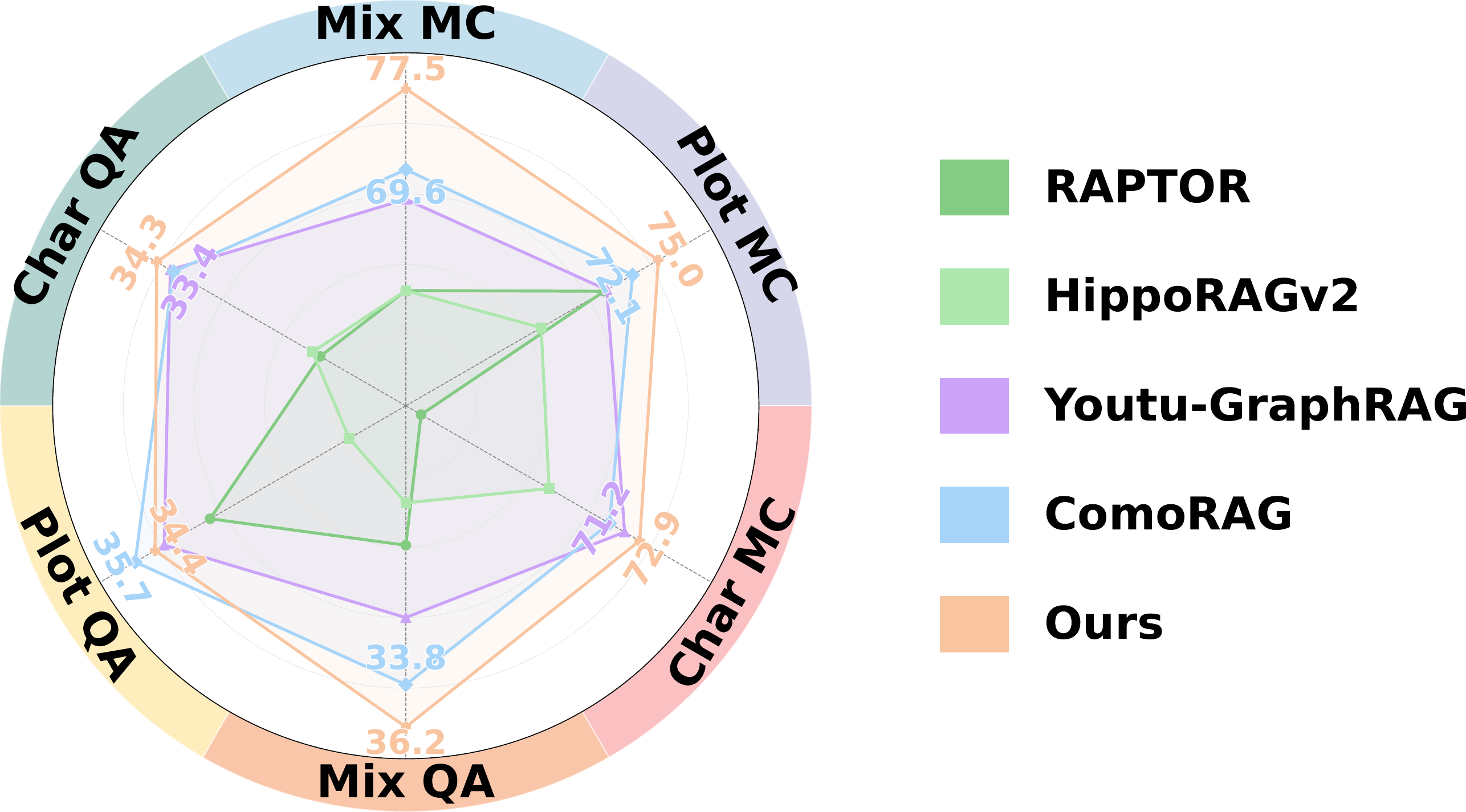}
\caption{Performance by query types of RAG methods. }
\label{fig:radia}
\vspace{-1.5ex}
\end{figure}
\subsection{Main Results}
\paragraph{Single-Step QA Performance.} From Table \ref{tab:single}, we conclude that: 1) Our framework outperforms all baselines across both QA and MC tasks. Notably, on MC tasks, it achieves a relative improvement of 11.56\% in average ACC compared to the strongest baseline. This gain demonstrates the robustness and efficacy of our proposed Pons bridging framework in synthesizing fragmented evidence. 2) Remarkably, PonsRAG achieves its largest performance gains on EN.MC, surpassing the second-best method by a margin of over 10\%. We attribute this to the longer length documents in EN.MC (150k+ tokens), which exacerbates the cognitive island. Crucially, as this cognitive gap widens, our framework demonstrates an increasing advantage by bridging these isolated islands detailed in Section \ref{sec:island}.
\paragraph{Multi-Step QA Performance.} Table \ref{tab:multi} demonstrates that: 1) PonsRAG achieves the highest performance across both QA and MC tasks. Notably, PonsRAG with IRCoT surpasses ComoRAG by over 5\% across all benchmarks. 2) The integration of iterative reasoning further unleashes the potential of our mechanisms as the NoCha Improv. increasing from 5.82\% to 13.75\%. We attribute this boost to the synergy between IRCoT and our architecture. The query rewriting mechanism in IRCoT generates new queries which anchors diverse nodes in Query Anchor stage. With these newly found anchor nodes, Pons Layer uncover hidden nodes that remain dormant under a single static query.
\begin{table}[t]
\centering
\small
\begin{tabular*}{0.9\columnwidth}{@{\extracolsep{\fill}}lccc} 
\toprule 
~~~~Method & \multicolumn{1}{c}{EN.MC} & \multicolumn{2}{c}{EN.QA} \\ 
\cmidrule(lr){2-2}\cmidrule(lr){3-4} 
& ACC & F1 & EM \\
\midrule 
~~~~\textbf{PonsRAG} & \textbf{77.73} & \textbf{35.13} & \textbf{26.21} \\ 
\midrule 
\multicolumn{4}{c}{\cellcolor{lightgray}\textit{\textbf{Index}}} \\ 
~~~~w/~~~~Char & 52.40 & 21.73 & 17.95\\ 
~~~~w/~~~~Plot & 55.02 & 23.37 & 18.23 \\ 
~~~~w/o ~Pons & 61.13 & 28.59 & 19.37 \\ 
\multicolumn{4}{c}{\cellcolor{lightgray}\textit{\textbf{Retrieval}}} \\ 
~~~~w/o Pons Awaken & 65.50 & 29.52 & 24.22 \\ 
~~~~w/o Pons Match & 64.19 & 30.90 & 21.65 \\
~~~~w/o Flow Filter & 70.74 & 32.93 & 25.07 \\
\bottomrule 
\end{tabular*} 
\caption{Ablation studies of PonsRAG.}
\label{tab:ablation_study} 
\vspace{-1.5ex}
\end{table}
\subsection{Ablation Studies}
\paragraph{Impact of Triple-layer Knowledge Source.} 
The three rows under Index in Table \ref{tab:ablation_study} detail the ablation of our knowledge source (w/ Char, w/ Plot, w/o Pons), revealing several key insights: 1) Relying on either the Char Layer or the Plot Layer yields suboptimal performance, as these single-layer configurations capture only fragmented narrative facts. 2) More importantly, a naive combination of these two layers without the connective Pons Layer remains constrained by the cognitive island problem. The inability to share evidence across these two layers disrupts coordinated reasoning resulting in a performance degradation with ACC dropping by approximately 20\% on EN.MC.
\paragraph{Effectiveness of Coordinated Reasoning.} To further ablate the coordinated reasoning pipeline, we remove each step except the Query Anchor as it initiates the entire pipeline. The results in Table \ref{tab:ablation_study} (three rows under Retrieval) show that each stage is essential, with performance decreasing at varying levels with each removal. Remarkably, the greatest impact is observed when removing Pons Match with ACC dropping by over 10\% on EN.MC. We find that this drop is caused by the influence of the main character. In this experiment, we modify the Hungarian Algorithm to select top 30 pairs with the highest edge weight, which leads to a phenomenon where a character matches multiple events. This directly causes the loss of evidence about the character, leading to the performance degradation. We further conduct experiments about the performance with different matching strategy in Appendix \ref{sec:island}.
\subsection{Detailed Analysis}\label{sec:island}
\begin{figure*}[t]
  \centering
  \begin{subfigure}{0.47\textwidth}
    \centering
    \includegraphics[width=\linewidth]{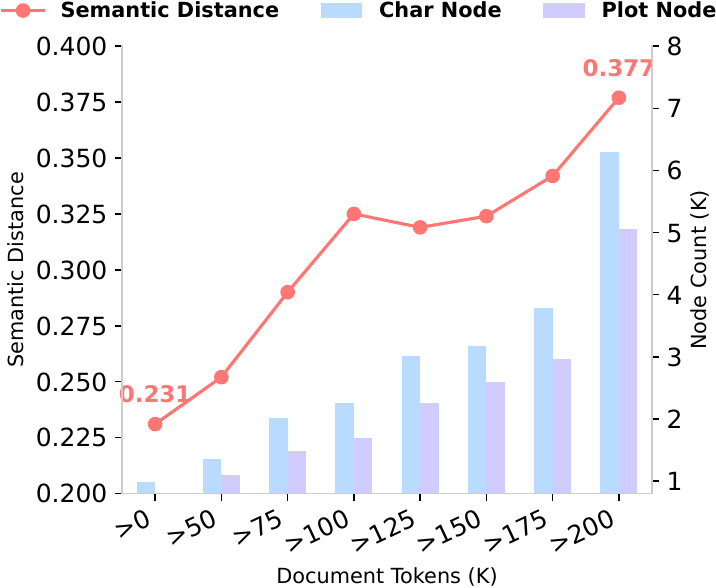}
    \caption{Semantic Distance varying along with document length.}
    \label{fig:id}
  \end{subfigure}%
  \hfill
  \begin{subfigure}{0.49\textwidth}
    \centering
    \includegraphics[width=\linewidth]{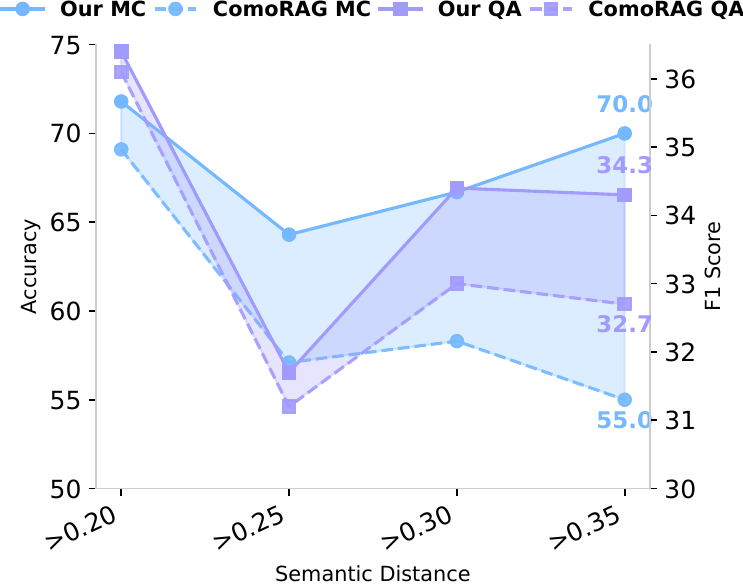}
    \caption{Performance gain along with Semantic Distance.}
    \label{fig:pg}
  \end{subfigure}

  \vspace{-1ex}
  \caption{Analysis of island density and performance gains.}
  \label{fig:id_total}
  \vspace{-1.5ex}
\end{figure*}

\paragraph{Longer Document Greater Separation.} 
Inspired by cluster separation theory \citep{lance1967general}, we use the average cross-layer semantic distance between character nodes and plot nodes as an indicator of cross-layer separation associated with cognitive island. Specifically, for a character node $u \in \mathcal{V}^{\text{char}}$ and a plot node $v \in \mathcal{V}^{\text{plot}}$, we define their semantic distance as $dis(u,v)=1-\cos(d_u,r_v)$, where $d_u$ denotes the description of character $u$, $r_v$ denotes the event details of the plot node $v$, and $\cos(\cdot,\cdot)$ denotes cosine similarity. Hence, the average cross-layer distance is then defined as:
\begin{equation}\label{eq:is}
D_{\text{cross}} =
\frac{\sum_{u \in \mathcal{V}^{\text{char}}} \sum_{v \in \mathcal{V}^{\text{plot}}} dis(u,v)}
{|\mathcal{V}^{\text{char}}| \cdot |\mathcal{V}^{\text{plot}}|}
\end{equation}
As illustrated in Figure~\ref{fig:id}, $D_{\text{cross}}$ rises from 0.231 to 0.377 as document length increases, suggesting that longer documents tend to exhibit greater cross-layer semantic separation between character and plot information.

\begin{table}[t]
\centering
\small

\setlength{\tabcolsep}{16pt} 
\begin{tabular*}{0.9\columnwidth}{@{\hspace{1em}\extracolsep{\fill}}ccc@{\hspace{1em}}}
\toprule
\textbf{$\tau$} & \textbf{EN.MC (ACC)} & \textbf{EN.QA (F1)} \\
\midrule
0.00 & $72.73 \pm 0.51$ & $31.75 \pm 0.59$ \\
0.25 & $73.62 \pm 0.34$ & $32.58 \pm 0.41$ \\
\midrule
0.50 & $77.31 \pm 0.21$ & $\textbf{34.61} \pm 0.34$ \\
0.75 & $\textbf{77.60} \pm \textbf{0.11}$ & $34.03 \pm \textbf{0.20}$ \\
\midrule
0.80 & $74.63 \pm 0.12$ & $33.12 \pm 0.17$ \\
0.90 & $74.20 \pm 0.09$ & $32.35 \pm 0.13$ \\
\bottomrule
\end{tabular*}
\caption{Performance under diverse Sparsity Controller $\tau$ (details in Equation~(\ref{eq:wvu})).}
\label{tab:tau}
\vspace{-1.5ex}
\end{table}
\paragraph{Greater Separation Greater Gains.} To validate the effectiveness of our triple-layer indexing in mitigating cross-layer separation, we compare PonsRAG with a strong baseline under varying levels of $D_{\text{cross}}$. As shown in Figure~\ref{fig:pg}, the performance gap consistently widens as $D_{\text{cross}}$ increases. Specifically, as $D_{\text{cross}}$ increases from 0.30 to 0.35, our ACC improves to 70\%, whereas ComoRAG drops to 55\%, widening the performance gap to 15\%. This divergence shows that, under severe cognitive islands, conventional frameworks struggle to connect fragmented evidence stored in distinct layers, whereas the Pons Layer in our index explicitly bridges all layers and turns structural complexity into a retrieval advantage.
\paragraph{Analysis of Pons Edge.} 
We further analyze the Pons Layer, the core component of our index, with a particular focus on the pons edges it introduces. To determine the key hyperparameter, the Sparsity Controller $\tau$ in Equation~(\ref{eq:wvu}), we examine how performance varies with $\tau$, as shown in Table~\ref{tab:tau}. The results reveal two insights: 1) As $\tau$ increases, the variance decreases from 0.51 to 0.09, indicating that performance becomes more stable at higher $\tau$ values. 2) Although the best performance is achieved under different settings for different tasks (i.e., $\tau = 0.75$ for MC and $\tau = 0.50$ for QA), performance drops sharply when $\tau < 0.50$ and $\tau > 0.75$. By contrast, when $\tau $ is between 0.50 and 0.75, the performance remains steady, with only a 0.58 difference in F1. Thus, the results suggest that the optimal $\tau$ lies between 0.50 and 0.75, as this range filters out noisy Pons edges while preserving informative ones. We further study the quality of these pons edges in Appendix \ref{sec:edge_eval} and other vital hyperparameters in Appendix \ref{app:mgs}.

\paragraph{Analysis of Query Resolution.} 
To better understand where our method yields the greatest benefit, we categorize all questions from the EN.MC and EN.QA datasets into three types (details about the classification method and statics of each query type are provided in Appendix~\ref{sec:appendix_query_classification}): 
\begin{itemize}[noitemsep,nolistsep,leftmargin=*]
    \item \textbf{Char Queries}: Queries centering on character traits or background details, e.g., \textit{``What religion is Octavio Amber?''}
    \item \textbf{Plot Queries}: Queries demanding narrative events along the plot line, e.g., \textit{``Where does Trace choose to live at the end of the novel?''}
    \item \textbf{Mix Queries}: Queries necessitating an understanding of both character traits and narrative events, e.g., \textit{``Who is the half crazed man named Arthur who worked with Norbert before Becky?''}
\end{itemize}
Based on this classification, we compare the performance of PonsRAG and the baseline on each query type. Results in Figure~\ref{fig:radia} show that the advantage of PonsRAG is most pronounced on Mix queries. Although ComoRAG achieves a 4\% lead over PonsRAG on Plot QA, it falls nearly 8\% behind on Mix QA. This gap suggests that our coordinated pipeline helps organize character and plot evidence more effectively across layers. To further illustrate the behavior of our coordinated reasoning pipeline, we provide a case study in Appendix~\ref{sec:goodcase}.
\paragraph{Analysis of Matching Constraints}
Narratives inherently feature many-to-many relationships. Therefore, we ablate the strict 1-to-1 constraint in Pons Match to test whether a relaxed 1-to-$N$ matching improves reasoning. We compare our approach ($N=1$) against $N=2$, $N=3$, and a Dense setting (retaining all edges in $\mathcal{G}_{sub}$ without pruning).
\begin{table}[h]
\centering
\resizebox{\columnwidth}{!}{
\small
\begin{tabular}{lccc}
\toprule
\textbf{Matching Strategy} & \textbf{EN.MC (ACC)} & \textbf{EN.QA (F1)} & \textbf{Avg Context Tokens} \\
\midrule
Dense (w/o Match) & 68.12 & 28.65 & 5,203 \\
1-to-3 & 73.03 & 30.34 & 3,856 \\
1-to-2 & 74.24 & 32.28 & 2,438 \\
\textbf{1-to-1 (Ours)} & \textbf{77.73} & \textbf{35.13} & \textbf{1,187} \\
\bottomrule
\end{tabular}}
\caption{Ablation on maximum degree constraints in Pons Match.}
\label{tab:match_ablation}
\end{table}
As shown in Table \ref{tab:match_ablation}, relaxing the mapping to 1-to-$N$ ($N \ge 2$) consistently degrades accuracy while substantially inflating the context token load. This confirms that the preceding \textbf{Pons Awaken} phase (via Co-HITS) already captures sufficient many-to-many semantic resonance. Consequently, \textbf{Pons Match} acts as a crucial sparsity regularizer rather than a recall expander. Routing a denser 1-to-$N$ graph to the \textbf{Flow Filter} floods the LLM with cross-layer noise and redundant tokens, severely exacerbating the "lost-in-the-middle" effect.
\section{Conclusion}
To address the cognitive-island failure mode in long narrative reasoning, we propose PonsRAG, a triple-layer retrieval framework that coordinates Character, Plot, and Pons layers for cross-layer evidence selection. PonsRAG delivers strong performance across four benchmarks, with advantages becoming more evident on longer documents, highlighting the usefulness of structured cross-layer retrieval in long-context settings.
\section*{Limitations}
Despite its strong performance on long narrative reasoning, PonsRAG still has limitations. Since the framework is explicitly designed for long-context, our evaluation is currently limited to LNR benchmarks. We have not yet examined its effectiveness on other reasoning settings, such as multi-hop QA or more general long-context tasks. Extending the coordinated reasoning paradigm to broader range of reasoning tasks remains an important direction for future work. 
\section*{Acknowledgments}
This paper was supported by the National Natural Science Foundation of China (No. 62306112), Guangdong Basic and Applied Basic Research Foundation (No. 2026A1515010253), and Guangdong S\&T Programme Key-Area Research and Development Program of Guangdong Province (2026B0101100004), National Natural Science Foundation of China (62276279), Guangdong Basic and Applied Basic Research Foundation (2024B1515020032).
\nocite{langley00}

\bibliographystyle{acl_natbib}
\bibliography{refs}

\newpage
\appendix





\section{Benchmark Details}
\label{sec:implementation details}
In this section, we detail the evaluated benchmarks and summarize their overall statistics as shown in Table \ref{tab:token_counts}.

\begin{table}[ht]
\centering
\resizebox{\columnwidth}{!}{
\small
\setlength{\tabcolsep}{4pt}
\begin{tabular}{lcccc}
\toprule
\textbf{Dataset} & \textbf{\#Docs} & \textbf{\#Queries} & \textbf{Total Tokens} & \textbf{Avg. Tokens} \\
\midrule
NarrativeQA & 17 & 500 & 887,763 & 52,221 \\
EN.QA       & 69 & 351 & 14,497,149 & 210,104 \\
EN.MC       & 58 & 229 & 11,249,572 & 193,958 \\
NoCha       & 4  & 126 & 555,972 & 138,993 \\
\bottomrule
\end{tabular}}
\caption{Statistics of benchmarks. \textbf{Avg. Tokens} denotes the average length of documents.}
\label{tab:token_counts}
\end{table}

\section{Setup and Hyperparameters}
\label{sec:Hyperparameters}
\subsection{Implementation Details}
For fairness, all RAGs use GPT-4o-mini as the backbone with temperature as 0.8 and a 6K-token context limit. All the structured and multi-step RAGs apply BGE-M3 as the embedding model with a 512-token chunk size.
\subsection{Hyperparameters Details}
For hyperparameters of PonsRAG, we follow HippoRAGv2 to construct the character knowledge graph in Char Layer and we set the sparsity controller $\tau$ to 0.75 for MC tasks and 0.50 for QA tasks in Pons Layer. To prevent data leakage, all hyperparameters (including $\tau$ and MGS weights) were exclusively optimized on a validation set. More detailed hyperparameters are shown in the Table \ref{tab:hyperparameters}
\begin{table}[h]
\centering
\small
\begin{tabular}{lc}
\toprule
\textbf{Hyperparameter} & \textbf{Value} \\
\midrule
\multicolumn{2}{l}{\textit{Model Selection}} \\
LLM Backbone Agent & GPT-4o-mini \\
Retrieval Embedding Model & BGE-M3 \\
\midrule
\multicolumn{2}{l}{\textit{Triple-Layer Indexing Setup}} \\
Max Chunk Size & 512 tokens \\
Sparsity Controller ($\tau$) & 0.75 (MC) / 0.50 (QA) \\
\midrule
\multicolumn{2}{l}{\textit{Query Anchor Phase}} \\
Anchor Top-$K$ ($k_1, k_2$) & 15 \\ 
MGS Event Weight ($\alpha$) & 0.7 \\ 
MGS Summary Weight ($\beta$) & 0.2 \\ 
MGS Chunk Weight ($\gamma$) & 0.1 \\ 
\midrule
\multicolumn{2}{l}{\textit{Pons Awaken \& Match Phase}} \\
Awaken Top-$K$ ($k_3$) & 15 \\ 
Match Top-$K$ ($k_4$) & 30 \\ 
\midrule
\multicolumn{2}{l}{\textit{Context Constraints}} \\
Max Context Length & 6,000 tokens \\
\bottomrule
\end{tabular}
\caption{Detailed hyperparameter settings of PonsRAG.}
\label{tab:hyperparameters}
\end{table}

\subsection{Analysis of MGS Weights}\label{app:mgs}

To analyse the MGS weights detailed in Equation (\ref{eq:scr}), we first conduct an ablation study of each weight on EN.MC and EN.QA with results are shown in three rows above the line in Table \ref{tab:hyperparam_results}. The results demonstrate that the weight of event $r_v$ contributes most to the performance as it provides the base details about a plot node whereas $s_v$ and $c_v$ can be shared across multiple nodes, potentially introducing noise that degrades performance. 

Therefore, we constrain $\alpha$ to be no smaller than $\beta$ and $\gamma$, and use a grid search to identify the best parameter setting. We report representative performance variations under different MGS weight settings in the bottom five rows of Table \ref{tab:hyperparam_results}, which suggest two main observations: 1) On both MC and QA tasks, PonsRAG consistently peaks at the setting of $\alpha=0.7, \beta=0.2, \gamma=0.1$. 2) Our performance is relatively insensitive to $\beta$ and $\gamma$ weights, with the ACC varying from 74.02 to 77.51, all of which surpass the second-best baseline. 

\begin{table}[ht]
\centering
\resizebox{\columnwidth}{!}{
\small
\setlength{\tabcolsep}{8pt} 
\begin{tabular}{lcc}
\toprule
\textbf{MGS Configuration} & \textbf{EN.MC (ACC)} & \textbf{EN.QA (F1)} \\
\midrule

$\alpha=1.0, \beta=0.0, \gamma=0.0$ & $74.02 \pm 0.42$ & $33.95 \pm 0.55$ \\
$\alpha=0.0, \beta=1.0, \gamma=0.0$ & $69.53 \pm 0.48$ & $32.08 \pm 0.49$ \\
$\alpha=0.0, \beta=0.0, \gamma=1.0$ & $67.74 \pm 0.39$ & $31.15 \pm 0.51$ \\
\midrule

$\alpha=0.5, \beta=0.1, \gamma=0.4$ & $74.25 \pm 0.65$ & $33.36 \pm 0.71$ \\
$\alpha=0.5, \beta=0.2, \gamma=0.3$ & $74.89 \pm 0.48$ & $34.01 \pm 0.54$ \\
$\alpha=0.6, \beta=0.1, \gamma=0.3$ & $75.78 \pm 0.27$ & $34.67 \pm 0.32$ \\
$\alpha=0.6, \beta=0.2, \gamma=0.2$ & $76.22 \pm 0.38$ & $34.35 \pm 0.48$ \\
$\alpha=0.7, \beta=0.2, \gamma=0.1$ & $\textbf{77.51} \pm \textbf{0.21}$ & $\textbf{35.13} \pm \textbf{0.25}$ \\
\bottomrule
\end{tabular}}
\caption{Performance of PonsRAG with different MGS configurations at $\tau = 0.75$ for EN.MC and $\tau = 0.50$ for EN.QA.}
\label{tab:hyperparam_results}
\end{table}

\section{Detailed Validation of Bridge Quality}
\label{sec:edge_eval}

To further justify the design of the Pons weight $w_{uv}$ in Equation \ref{eq:wvu}, we compare our method against two common alternative bridging strategies:
\begin{itemize}[leftmargin=*,itemsep=0pt]
    \item \textbf{Entity Mention}: A link is established only if the character's name explicitly appears in the event chunk.
    \item \textbf{Pure Semantic}: Edges are weighted solely by $sim(d_u, s_v)$ without the frequency-balancing term $\mathcal{I}_u$
\end{itemize}

As shown in Table \ref{tab:edge_quality}, while Entity Mention achieves high precision, it fails to recover latent narrative links, leading to suboptimal downstream performance. Pure Semantic retrieval suffers from noise introduced by high-frequency characters. Our Pons Weight balances these factors, providing the most effective context for long narrative reasoning.
\begin{table}[ht]
\centering
\small 
\setlength{\tabcolsep}{4pt} 
\begin{tabularx}{\columnwidth}{lccc} 
\toprule
\textbf{Bridging Scheme} & \textbf{Prec@100} & \textbf{EN.MC} & \textbf{EN.QA} \\
& & \textbf{(ACC)} & \textbf{(F1)} \\  
\midrule
Entity Mention & 91.0\% & 58.45 & 24.82 \\
Pure Semantic  & 64.0\% & 68.12 & 30.95 \\
\textbf{Pons Weight (Ours)} & \textbf{82.0\%} & \textbf{75.54} & \textbf{34.56} \\
\bottomrule
\end{tabularx}
\caption{Manual precision of edges and downstream performance on EN.MC and EN.QA across different bridging schemes. Downstream results for Ours are consistent with Table \ref{tab:ablation_study}.} 
\label{tab:edge_quality}
\end{table}

\section{Query Type Details}
\label{sec:appendix_query_classification}

\paragraph{Classification Protocol.} 
To systematically diagnose the reasoning bottlenecks (as discussed in Section \ref{sec:island}.), we employ GPT-4o as an automated annotator to categorize queries into three distinct types: \textbf{Char}, \textbf{Plot}, and \textbf{Mix}. To quantify the reliability of these automated labels, two human experts independently annotated a random sample of 100 queries. The specific instruction template used for this automated classification is detailed in Appendix~\ref{sec:Query Type Classification Prompt}. 

\paragraph{Query Distribution.}
As summarized in Table~\ref{tab:query_distribution}, applying this pipeline to EN.MC and EN.QA yields approximately 30\% Char, 26\% Plot, and 44\% Mix queries. This confirms that the benchmarks heavily emphasize complex joint reasoning while retaining adequate single-aspect coverage.

\begin{table}[htbp]
\centering
\small
\begin{tabular}{lcccc}
\toprule
\textbf{Query Type} & \textbf{EN.MC} & \textbf{EN.QA} & \textbf{Total} & \textbf{Proportion} \\
\midrule
\textbf{Char} & 59 & 114 & 173 & 29.8\% \\
\textbf{Plot} & 68 & 86 & 154 & 26.6\% \\
\textbf{Mix} & 102 & 151 & 253 & 43.6\% \\
\midrule
\textbf{Total} & 229 & 351 & 580 & 100.0\% \\
\bottomrule
\end{tabular}
\caption{Detailed distribution and counts of query types across the EN.MC and EN.QA datasets.}
\label{tab:query_distribution}
\end{table}

\paragraph{Annotation Reliability.} 
 As shown in Table~\ref{tab:confusion_matrix}, the automated annotator achieved an 88\% accuracy against the human consensus, yielding a Cohen's $\kappa$ of 0.81 (which indicates strong agreement). The confusion matrix reveals that the primary source of discrepancy lies in distinguishing complex Plot queries from Mix queries, since tracking multi-step plot events can sometimes implicitly necessitate character-level reasoning. Nevertheless, the overall misclassification rate remains sufficiently low, ensuring that our mechanistic observations in Figure~4 are statistically robust.

\begin{table}[htbp]
\centering
\small
\begin{tabular}{lcccc}
\toprule
& \multicolumn{3}{c}{\textbf{Predicted (GPT-4o)}} \\
\cmidrule{2-4}
\textbf{True (Human)} & \textbf{Char} & \textbf{Plot} & \textbf{Mix} & \textbf{Total} \\
\midrule
\textbf{Char} & 25 & 3 & 1 & 29 \\
\textbf{Plot} & 1 & 22 & 3 & 26 \\
\textbf{Mix} & 0 & 4 & 41 & 45 \\
\midrule
\textbf{Accuracy} & \multicolumn{4}{c}{\textbf{88.0\%}} \\
\bottomrule
\end{tabular}
\caption{Confusion matrix of LLM-based query classification against human annotation on 100 sampled queries.}
\label{tab:confusion_matrix}
\end{table}

\section{Discussion on GraphRAG and HiRAG}
\label{sec:appendix_excluded_baselines}
In this section, we compare PonsRAG with GraphRAG and HiRAG on a subset of EN.QA. GraphRAG constructs a graph-structured knowledge index and retrieves evidence over entities and their relations. HiRAG adopts a coarse-to-fine retrieval strategy, progressively narrowing from global context to specific details for reasoning. Both methods have shown strong performance on multi-hop reasoning tasks. 

However, as shown in Table~\ref{tab:cost_comparison}, they struggle with long narrative reasoning, exhibiting both substantially lower performance and much higher cost. In particular, HiRAG achieves only 21.37 (F1), roughly \textbf{60\%} of our performance, while incurring about \textbf{80} $\times$ higher token costs and \textbf{8} $\times$ higher time costs. Considering this unfavorable cost--performance trade-off, we exclude them from the remaining benchmarks.

\begin{table}[htbp]
\centering
\resizebox{\columnwidth}{!}{
\footnotesize 
\setlength{\tabcolsep}{2.5pt} 
\begin{tabular}{lccc}
\toprule
\textbf{Metrics} & \textbf{PonsRAG} & \textbf{GraphRAG} & \textbf{HiRAG} \\
\midrule
\multicolumn{4}{l}{\textit{Performance}} \\
\quad F1 Score & 34.38 (100\%) & 14.60 (42.5\%) & 21.37 (62.2\%) \\
\quad EM Score & 25.31 (100\%) & 8.20 (32.4\%) & 14.28 (56.4\%)\\
\midrule
\multicolumn{4}{l}{\textit{Token Usage}} \\
\quad Tokens & 1.08M (100\%) & 26.43M (2447\%) & 95.81M (8871\%)\\ 
\midrule
\multicolumn{4}{l}{\textit{Average Time (s)}} \\ 
\quad Index & 608 (100\%) & 1936 (318.4\%) & 4763 (783.4\%) \\
\quad Retrieve & 6 (100\%) & 29 (483.3\%) & 49 (816.7\%) \\
\bottomrule
\end{tabular}}
\caption{Comparison of performance, token usage, and latency across different RAG paradigms. Percentages indicate the relative ratio compared to PonsRAG.}
\label{tab:cost_comparison}
\end{table}

\onecolumn
\section{Gold Case} \label{sec:goodcase}

To further illustrate the behavior of PonsRAG, Table \ref{tab:case_study} presents a specific case study. Given an incoming query $q$: \textit{``How does Jimmy Doyle spend all of his money with his friends in After the Race?''}, existing methods primarily retrieve surface-level event nodes (e.g., \textit{Business Investment}), which often mislead the LLM. In contrast, PonsRAG leverages the character anchor nodes $u \in \mathcal{V}_{anc}^{char}$ (\textit{Jimmy} and \textit{Routh}) initialized during the Query Anchor phase to uncover the latent key evidence $v \in \mathcal{V}_{awk}^{plot}$ (\textit{Card Games}) during the Pons Awaken stage. Subsequently, valid cross-layer connections are formalized into an optimal matching set $\mathcal{P}_{match}$ during the Pons Match phase. Finally, the Flow Filter mechanism prunes distracting noise to reconstruct a coherent chronological storyline $\mathcal{S}_{final}$. This refined context provides the precise evidential support necessary for the Generator to derive the correct answer $\hat{A}$.

\begin{table*}[!htbp]
\centering
\small
\begin{tabular}{>{\raggedright\arraybackslash}p{0.98\textwidth}}
\toprule
\multicolumn{1}{c}{\textbf{Input Data (No Options)}} \\
\textbf{Query:} How does Jimmy Doyle spend all of his money with his friends in “After the Race”? \\
\textbf{Options:}
[A] Betting on car racing
[B] Playing cards
[C] Arm wrestling
[D] Treating his friends to drinks
\\
\midrule
\multicolumn{1}{c}{\textbf{PonsRAG's Choice Result}} \\
\midrule
\multicolumn{1}{c}{\cellcolor{lightgrey}\textit{\textbf{Query Anchor}}}\\
\textbf{Anchor Char Nodes $\mathcal{V}_{anc}^{char}$:} \\ 
- \sethlcolor{lightgreen}\hl{JIMMY (PERSON)}: Jimmy is the son of a wealthy butcher, educated in England and known for his popularity and social life...\\
- \sethlcolor{lightgreen}\hl{ROUTH (PERSON)}: Routh is a young Englishman who is part of the dinner party...\\
...\\
\textbf{Anchor Plot Nodes $\mathcal{V}_{anc}^{plot}$:} \\
- \sethlcolor{lightgreen}\hl{Car Race}: The French cars, which had finished solidly in the race...\sethlcolor{lightgreen}\hl{Segouin is the owner of one of the cars}...\\
- \sethlcolor{lightgreen}\hl{Business Investment}: Jimmy thinks about \sethlcolor{lightgreen}\hl{a business investment in the motor industry}... believing it to be a good opportunity...\\
...\\
\midrule
\multicolumn{1}{c}{\cellcolor{lightgrey}\textit{\textbf{Pons Awaken}}}\\
\textbf{Awaken Char Nodes $\mathcal{V}_{awk}^{char}$:} \\
- \sethlcolor{lightpurple}\hl{SEGOUIN (PERSON)}: Segouin is an acquaintance of Jimmy, reputed to own hotels in France...\\
- \sethlcolor{lightpurple}\hl{FARRINGTON (PERSON)}: Farrington is an employee who is being reprimanded by Mr Alleyne...\\
...\\
\textbf{Plot Nodes $\mathcal{V}_{awk}^{plot}$:} \\
- \sethlcolor{lightpurple}\hl{Arm Wrestling}: \sethlcolor{lightpurple}\hl{Weathers defeats Farrington in a hand wrestling match},...\\
- \sethlcolor{lightpurple}\hl{Card Games}:    Jimmy struggles with card games...\sethlcolor{lightpurple}\hl{frequently mistook his cards, leading to confusion and losses}...Routh...ultimately winning amidst the excitement of the gathering....\\
...\\
\midrule
\multicolumn{1}{c}{\cellcolor{lightgrey}\textit{\textbf{Pons Match}}}\\
\textbf{Matching Set $\mathcal{P}_{match}$:}\\
- \sethlcolor{lightgreen}\hl{JIMMY...} $ \iff $ \sethlcolor{lightgreen}\hl{Business Investment...}\\
- \sethlcolor{lightgreen}\hl{ROUTH...} $ \iff $ \sethlcolor{lightpurple}\hl{Card Games...}\\
- \sethlcolor{lightpurple}\hl{FARRINGTON...} $ \iff $ \sethlcolor{lightpurple}\hl{Arm Wrestling...}\\
- \sethlcolor{lightpurple}\hl{SEGOUIN...} $ \iff $ \sethlcolor{lightgreen}\hl{Car Race...}\\
\midrule
\multicolumn{1}{c}{\cellcolor{lightgrey}\textit{\textbf{Flow Filter}}}\\
\textbf{Matching Sequence $\mathcal{S}_{match}$:}\\
$ \langle \text{JIMMY}, \text{Business Investment} \rangle $ 
$ \Longrightarrow $ 
$ \langle \text{SEGOUIN}, \text{Car Race} \rangle $
$ \Longrightarrow $ 
...
$ \Longrightarrow $ 
$ \langle \text{FARRINGTON}, \text{Arm Wrestling} \rangle $
$ \Longrightarrow $ 
...
$ \Longrightarrow $ 
$ \langle \text{ROUTH}, \text{Card Games} \rangle $
$ \Longrightarrow $ 
...\\
\textbf{Final Sequence $S_{final}$:}\\
$ \langle \text{SEGOUIN}, \text{Car Race} \rangle $
$ \Longrightarrow $ 
...
$ \Longrightarrow $ 
$ \langle \text{ROUTH}, \text{Card Games} \rangle $\\
\midrule
\textbf{Chosen:} \textcolor{green}{\textbf{B.}(Correct)}\\
(B) Playing cards: The text explicitly states that Jimmy participates in card games, frequently mistaking his cards and ultimately losing money.\\
\bottomrule
\end{tabular}
\caption{\textbf{Case Study on Coordinated Narrative Reasoning.} We present a case to demonstrate our model's performance in long-context understanding. Different colors are used to highlight the nature of the processed information: \sethlcolor{lightgreen}\hl{Green} is used for key cognition found in step \textbf{Query Anchor} that contributes to the correct answer, while \sethlcolor{lightpurple}\hl{Purple} is used for the evidence in step \textbf{Pons Awaken}.}
\label{tab:case_study}
\end{table*}

\section{Prompting Templates}
\label{sec:appendix_prompts}

Below are the detailed instruction templates utilized across the different modules of our framework.

\definecolor{promptgreen}{HTML}{50A050} 

\newtcolorbox{promptbox}[2][]{
    enhanced,
    breakable,                   
    colback=promptgreen!5!white, 
    colframe=promptgreen,        
    coltitle=white,              
    fonttitle=\bfseries,         
    title={\tcbox[colback=white, colframe=promptgreen, coltext=promptgreen, size=fbox, boxsep=1pt, boxrule=0.5pt, arc=1pt, on line]{\textsf{\small #1}} \hspace{2mm} #2},
    arc=2pt,                     
    boxrule=1pt,                
    left=4pt, right=4pt, top=2pt, bottom=2pt, 
    before skip=16pt,            
    after skip=16pt              
}

\begin{promptbox}[Char Layer]{Instruction Template for Entity and Description Extraction}
\textbf{Role}\\
Your task is to extract entities of specific types from the given text.\\

\textbf{Task}\\
For each entity, identify:
\begin{enumerate}[leftmargin=*, noitemsep, topsep=2pt]
    \item \texttt{entity\_name}: capitalized name of the entity
    \item \texttt{entity\_type}: one of the provided types (or “normal\_entity” if none match)
    \item \texttt{entity\_description}: a concise description of the entity's attributes and activities
\end{enumerate}
\vspace{2mm}

\textbf{Response Format}\\
Return the result as a list of tuples in the following format:\\
\texttt{("entity"\#\#\#\allowbreak<entity\_name>\#\#\#\allowbreak<entity\_type>\#\#\#\allowbreak<entity\_description>@@@)}\\
End the response with \texttt{<end>}.\\
\end{promptbox}

\begin{promptbox}[Char Layer]{Instruction Template for Entity Description Summarization}
\textbf{Role}\\
You are a helpful assistant responsible for generating a comprehensive summary of the data provided below.\\

\textbf{Task}\\
Given one or two entities and a list of related descriptions, synthesize them into a single summary by following these rules:
\begin{enumerate}[leftmargin=*, noitemsep, topsep=2pt]
    \item Merge all provided descriptions into a single, comprehensive text, ensuring no collected information is omitted.
    \item If the provided descriptions conflict, logically resolve these contradictions to produce a coherent summary.
    \item Write strictly in the third person and explicitly include the entity names for full context.
\end{enumerate}
\vspace{2mm}

\textbf{Input Format}\\
\texttt{Entity: \$\{entity\}}\\
\texttt{Descriptions: \$\{descriptions\}}\\

\textbf{Response Format}\\
Provide a single, comprehensive description that synthesizes all the provided descriptions into a coherent summary.\\
\end{promptbox}

\begin{promptbox}[Plot Layer]{Instruction Template for Event Extraction}
\textbf{Role}\\
You are an expert event extraction assistant.\\

\textbf{Task}\\
Please read the following text carefully and extract all key events in chronological order. Make sure:
\begin{enumerate}[leftmargin=*, noitemsep, topsep=2pt]
    \item Include all major and minor events mentioned.
    \item Maintain chronological order.
    \item Each event description should be concise (no more than one sentence).
    \item Each detail should clearly explain what happened, who was involved.
    \item Do not add any commentary or analysis beyond the events themselves.
\end{enumerate}
\vspace{2mm}

\textbf{Input Format}\\
\texttt{Text to analyze: \$\{chunk\_text\}}\\

\textbf{Response Format}\\
Return the result as a list of tuples in the following format:\\
\texttt{(<event description>\#\#\#\allowbreak<event details>)@@@}\\
End the response with \texttt{<end>}.\\
\end{promptbox}

\begin{promptbox}[Plot Layer]{Instruction Template for Event Summarization}
\textbf{Role}\\
You are an expert event summarization assistant.\\

\textbf{Task}\\
Please read the following events in chronological order carefully and generate a comprehensive summary. Make sure:
\begin{enumerate}[leftmargin=*, noitemsep, topsep=2pt]
    \item Include all events in the summary.
    \item Include all details about each event.
\end{enumerate}
\vspace{2mm}
\textbf{Input Format}\\
\texttt{Events: \$\{events\}}\\

\textbf{Response Format}\\
Return the summary directly without any additional commentary or analysis.\\
\end{promptbox}

\begin{promptbox}[Flow Filter]{Instruction Template for Character-Event Pair Filtering}
\textbf{Role}\\
You are a critical component of a high-stakes question-answering system used by top researchers and decision-makers worldwide.\\

\textbf{Task}\\
1. Identify pairs that are helpful to answer the user's query, if none are helpful, output \texttt{NONE}.\\
2. Output the specific indices of the selected pairs.\\

\textbf{Input}\\
You are given a Question and several Pairs (Characters, Events) from an article.\\

\textbf{Response Format}\\
ONLY OUTPUT INDICES SEPARATED BY COMMA!!! \\
\textit{Example:} \texttt{0,3,5,7,8,10,14}\\

\textbf{Limits}
\begin{itemize}[leftmargin=*, noitemsep, topsep=2pt]
    \item The accuracy of your response is paramount, as it will directly impact the decisions made by these high-level stakeholders!!!
    \item You must only use character or events from the candidate list and do not generate new ones!!!
\end{itemize}
\end{promptbox}

\begin{promptbox}[QA Answering]{Instruction Template for Final Answer Generation}
\textbf{Role}\\
You are an expert at carefully reading complex texts, extracting narrative details, and making logical inferences.\\

\textbf{Task}\\
Given the following detail article from a book, and a related question, you need to provide a comprehensive and accurate answer based on the given information.\\

\textbf{Input}\\
The context comprises extracted character profiles and a chronologically ordered sequence of narrative events, supported by their original text chunks.\\
\texttt{Context: \{<role\_1, event\_1>\, <role\_2, event\_2>\...<role\_n, event\_n>\}}\\
\texttt{Question: \{question\}}\\

\textbf{Response Format}
\begin{itemize}[leftmargin=*, noitemsep, topsep=2pt]
    \item \textbf{Content Understanding}: Start with a brief summary of the content in no more than two sentences.\texttt{\#\#\# Content Understanding}
    \item \textbf{Relevant Information Analysis}: Provide a markdown list of all relevant evidence strictly from retrieved documents.\texttt{\#\#\# Relevant Information Analysis}
    \item \textbf{Key Facts}: List the key facts that directly support the answer.\texttt{\#\#\# Key Facts}
    \item \textbf{Final Answer}: Provide the shortest possible answer taken directly from the text.\texttt{\#\#\# Final Answer}
\end{itemize}
\vspace{2mm}

\textbf{Limits}
\begin{itemize}[leftmargin=*, noitemsep, topsep=2pt]
    \item Do not infer or assume anything not explicitly stated in the retrieved documents.
    \item Do not fabricate facts.
    \item Prefer answers supported by multiple independent pieces of evidence.
\end{itemize}
\end{promptbox}

\vspace{2mm}
\section{Prompt Template for Query Type Classification}
\label{sec:Query Type Classification Prompt}

\definecolor{promptblue}{HTML}{4A90E2}

\newtcolorbox{bluepromptbox}[2][]{
    enhanced,
    breakable,                 
    colback=promptblue!5!white, 
    colframe=promptblue,         
    coltitle=white,              
    fonttitle=\bfseries,       
    title={\tcbox[colback=white, colframe=promptblue, coltext=promptblue, size=fbox, boxsep=1pt, boxrule=0.5pt, arc=1pt, on line]{\textsf{\small #1}} \hspace{2mm} #2},
    arc=2pt,                
    boxrule=1pt,               
    left=4pt, right=4pt, top=2pt, bottom=2pt, 
    before skip=16pt,           
    after skip=16pt              
}

\begin{bluepromptbox}[Query Classification]{Three-shot Demonstration Template}
\textbf{Three-shot Demonstration:}\\

\texttt{\{"QUERY": "What religion is Octavio Amber?"\}}\\
\texttt{\{"ID": "char"\}}\\[4pt]

\texttt{\{"QUERY": "Where does Trace choose to live at the end of the novel?"\}}\\
\texttt{\{"ID": "plot"\}}\\[4pt]

\texttt{\{"QUERY": "Who is the half crazed man named Arthur who worked with Norbert before Becky?"\}}\\
\texttt{\{"ID": "mix"\}}\\

\end{bluepromptbox}

\begin{bluepromptbox}[Query Classification]{Instruction Template}
\textbf{Role}\\
You are an expert on evaluating and classifying query types.\\

\textbf{Task}
\begin{enumerate}[leftmargin=*, noitemsep, topsep=2pt]
    \item Understand and identify the information needed to answer the given query.
    \item Classify the query into three types based on the retrieval focus: 
    \begin{itemize}[leftmargin=*, noitemsep, topsep=2pt]
        \item\texttt{char}, which focuses on character profiles; 
        \item\texttt{plot}, which focuses on narrative events;  
        \item\texttt{mix}, which requires joint reasoning to synthesize relations between character and event.
        \end{itemize}
\end{enumerate}
\vspace{2mm}

\textbf{Input}\\
\texttt{\{"QUERY": "XXX"\}}
\vspace{2mm}

\textbf{Output}\\
\texttt{\{"ID": "XXX"\}}

\end{bluepromptbox}
\vspace{1mm}
\end{document}